\documentclass{ecai}
\usepackage{times}
\usepackage{graphicx}
\usepackage{latexsym}

\usepackage[ruled]{algorithm2e}
\usepackage{amsmath}
\usepackage{booktabs}
\usepackage{threeparttable}
\ecaisubmission   

\begin{document}

\title{Efficient Integer-Arithmetic-Only Convolutional Neural Networks}

\author{Hengrui Zhao \and Dong Liu \and Houqiang Li\institute{University of Science and Technology of China, email: zhenry@mail.ustc.edu.cn, dongeliu@ustc.edu.cn, lihq@ustc.edu.cn} }

\maketitle
\bibliographystyle{ecai}

\begin{abstract}
Integer-arithmetic-only networks have been demonstrated effective to reduce computational cost and to ensure cross-platform consistency. However, previous works usually report a decline in the inference accuracy when converting well-trained floating-point-number (FPN) networks into integer networks. We analyze this phonomenon and find that the decline is due to activation quantization. Specifically, when we replace conventional ReLU with Bounded ReLU, how to set the bound for each neuron is a key problem. Considering the tradeoff between activation quantization error and network learning ability, we set an empirical rule to tune the bound of each Bounded ReLU. We also design a mechanism to handle the cases of feature map addition and feature map concatenation. Based on the proposed method, our trained 8-bit integer ResNet outperforms the 8-bit networks of Google's TensorFlow and NVIDIA's TensorRT for image recognition. We also experiment on VDSR for image super-resolution and on VRCNN for compression artifact reduction, both of which serve for regression tasks that natively require high inference accuracy. Our integer networks achieve equivalent performance as the corresponding FPN networks, but have only 1/4 memory cost and run 2$\times$ faster on modern GPUs. Our code and models can be found at github.com/HengRuiZ/brelu.
\end{abstract}

\section{INTRODUCTION}
Convolutional neural networks (CNNs) have demonstrated state-of-the-art performance in a lot of computer vision and image processing tasks, including classification~\cite{NIPS2012_4824}, detection~\cite{Redmon_2016_CVPR}, image enhancement~\cite{Kim_2016_CVPR} and video coding~\cite{10.1007/978-3-319-51811-4_3}. However, there is still a major difficulty to deploy CNNs in practice, especially in scenarios where computing resource is limited, since the CNNs have a huge number of parameters and require much computation. Many researches have been conducted to address this problem, for example by designing computationally efficient networks, exemplified by MobileNet~\cite{DBLP:journals/corr/HowardZCKWWAA17}, SqueezeNet~\cite{DBLP:journals/corr/IandolaMAHDK16}, and ShuffleNet~\cite{Zhang_2018_CVPR}, or by pruning a well-trained network to reduce complexity~\cite{LeCun:1989:OBD:2969830.2969903} while maintaining accuracy~\cite{NIPS2015_5784}.

A distinctive approach to efficient CNNs is using low bit-depth for convolutions. In earlier years, networks with extremely low bit-depth, such as binary neural networks~\cite{NIPS2016_6573}, ternary weight networks~\cite{DBLP:journals/corr/LiL16,DBLP:journals/corr/ZhuHMD16,DBLP:journals/corr/MellempudiKM0KD17}, and XNOR-net~\cite{10.1007/978-3-319-46493-0_32} have been proposed. These networks do not catch up with the current trend of using deeper and wider network~\cite{2014arXiv1409.1556S,He_2016_CVPR,Huang_2017_CVPR}. Recently, new methods are studied such as multi-bit compression~\cite{1510.00149,DBLP:journals/corr/ZhouYGXC17,DBLP:journals/corr/GuptaAGN15}, vector quantization~\cite{DBLP:journals/corr/GongLYB14}, hashing trick~\cite{DBLP:journals/corr/ChenWTWC15}, and ADMM~\cite{DBLP:journals/corr/LengLZJ17}. Among them, integer-arithmetic-only inference~\cite{Jacob_2018_CVPR}, which quantizes a floating-point-number (FPN) network to integers, seems a good solution.
FPN arithmetic is not friendly to digital computing devices. It costs much more computing power than integer arithmetic.
In addition, there is no standard of FPN arithmetic, so the implementation of FPN arithmetic is platform dependent~\cite{Goldberg:1991:CSK:103162.103163}---this is a significant drawback as for applications concerning interoperability, such as video coding.
Integer networks provide the benefit of smaller model, faster inference, as well as cross-platform consistency.

In previous works, integer networks were always reported worse than the corresponding FPN networks in accuracy, if given the same number of parameters. According to~\cite{Jacob_2018_CVPR}, the 8-bit integer ResNet152 achieves 76.7\% accuracy, which is only slightly higher than the FPN ResNet50 (76.4\%). Likewise, NVIDIA's TensorRT reported 74.6\% accuracy of 8-bit ResNet152, which is even lower~\cite{migacz20178}. The accuracy decline of very deep neural network has been a severe discouraging fact of the previous integer networks, especially considering that the network complexity has been multiplied to achieve only marginal improvement.

We want to achieve integer networks that perform as well as FPN networks. For this purpose, we analyze previous quantization methods and observed that, as~\cite{migacz20178} have claimed, if we quantize the weights linearly into 8-bit but keep the other modules unchanged, then the accuracy can be as high as before quantization. However, if both weights and activation is quantized into 8-bit linearly, significant drop of accuracy occurred---it is due to the low-precision activations. Nonetheless, previous works do not address this issue well.

Different from parameter quantization that quantizes static weights and biases into integer, activation quantization is dynamic as it quantizes the computed activations into integer during network inference. There are two possibilities for activation quantization: the first is to decide the quantization step on-the-fly, and the second is to determine the quantization step during training. We prefer the second way because it not only saves online computation but also provide higher precision.
For the activation quantization, our key idea is to adapt the bounds of activation function to feature maps and networks. Bounded activation function has been studied like ReLU6~\cite{DBLP:journals/corr/abs-1801-04381}, but it is too simple for various networks and datasets. We introduce the Bounded ReLU (BReLU) as the activation function into CNNs. Different from the widely used ReLU, BReLU has both a lower bound and an upper bound and is linear between the two bounds. The bounds of BReLU are adjustable to suit for different feature maps. Note that there is a fundamental tradeoff for BReLU followed by quantization. If the dynamic range of BReLU is large, the quantization step should be large, too (given a predefined bit-depth for activations), which loses precision. But if the dynamic range is small, many features will be deactivated, which limits the learning capability of CNNs. We then propose methods to calculate the dynamic range for BReLU adaptively to the training data and networks to address this issue.

We have verified the proposed method on three different CNN-based tasks: image classification, image super-resolution, and compression artifact reduction, all achieving state-of-the-art performance. The following two tasks belong to regression that natively requires high inference accuracy. Previously, low-bit-depth networks are seldom evaluated for these regression tasks.
In addition, with the help of CUDA-enabled devices to perform fast short-integer convolutions, we manage to convert 32-bit FPN networks into integer-arithmetic-only networks with 8-bit weights and 7-bit activations. Our integer networks can still achieve virtually the same accuracy as FPN networks, but have only 1/4 memory cost and run 2$\times$ faster on modern GPUs.

\section{RELATED WORK}
The closest work to us is~\cite{Jacob_2018_CVPR}, which is also the built-in method in Google's TensorFlow. It applies exponential moving averages for the activation quantization, which calculates the bounds of activations on-the-fly. It modifies the bounds after each iteration, which is too frequent to be suitable for parameter learning. Moreover, the exponential moving averages method requires a traversal for each feature map, which is computationally expensive. Another limitation is that all kernels in a concatenation layer are required to have the same quantization. We propose a new method to decide the bounds of BReLU. And we develop a more delicate ratio synchronization mechanism to address the concatenation layer.

Another integer network method is NVIDIA's TensorRT~\cite{migacz20178}. It proposes a relative entropy method to determine the activation bounds. The idea is to minimize the loss of information during quantization.
However, minimizing the defined loss cannot ensure the preserving of inference accuracy.
In addition, TensorRT relies on floating-point arithmetic, i.e. the resulting network is not integer only.

There are many works on network compression~\cite{DBLP:journals/corr/abs-1805-06085,DBLP:journals/corr/abs-1709-01134,DBLP:journals/corr/abs-1809-04191,DBLP:journals/corr/ZhouNZWWZ16}, but they are based on floating-point scaling that is very different from integer arithmetic. Since they still depend on floating-point arithmetic, it needs more inference cost and cannot promise interoperability.

\section{METHOD}
\subsection{Architecture}
Given a pretrained 32-bit floating-point-number (float32) convolutional network, our goal is an integer-arithmetic-only network with low-precision integer weights and activations, which will decrease the model size and accelerate inference. Similar to~\cite{Jacob_2018_CVPR}, our integer networks use 8-bit integer (int8) for weights and activations. The convolution results in bigger numbers, after which we quantize the activations back to int8 numbers, which are used for further convolutions.

\begin{figure*}[h]
\centerline{\includegraphics[width=0.8\textwidth]{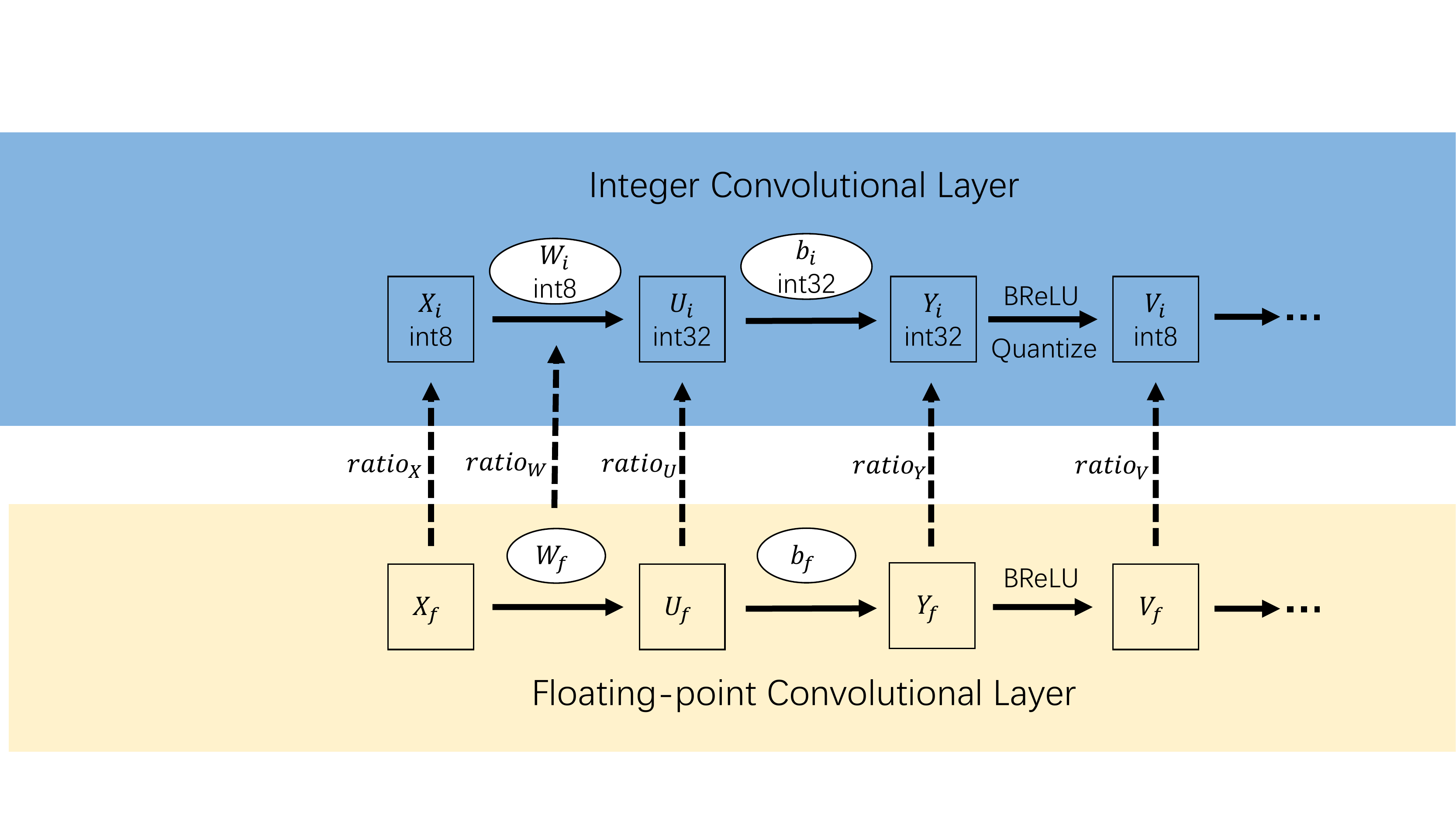}}
\caption{Illustration of the correspondence between an FPN convolutional layer and our integer-arithmetic-only convolutional layer, both equipped with our proposed BReLU.} \label{fig1}
\end{figure*}

As shown in Figure~\ref{fig1}, for an integer convolutional layer we perform integer convolution on input $X_i$ and weights $W_i$ to get feature maps $U_i$. $U_i$ is adjusted by the bias $b_i$ to get $Y_i$, which is then processed by the specific activation function--Bounded ReLU and quantized to get $V_i$. $V_i$ is the input of the next layer.

To achieve such an integer network, we quantize the weights and biases of a float32 network. For the quantization, we observed the parameters in a convolutional kernel $W_f$ roughly obey a normal distribution with zero mean, for which linearly quantization is efficient in precision and saves complexity compared to quantization with zero-shift. To fully utilize the limited precision of short integer, we map the maximal absolute value of $W_f$ to the maximal possible absolute value of integer. Specifically, for a given float32 convolutional kernel $W_f$ and int8 target,
\begin{equation}\label{quantizeW}
W_i=[\frac{W_f}{\Delta}]
\end{equation}
where $\Delta$ indicates the quantization step for $W_f$:
\begin{equation}
\Delta=\frac{\max(abs(W_f))}{127}
\end{equation}
To achieve the best accuracy, it is natural to make the integer networks close to the float32 networks during inference, because the float32 networks are well-trained. Accordingly, we assume a linear mapping between each of the kernels, feature maps and other parameters of an integer network and its floating-point counterpart:
\begin{equation}
X_i \approx X_f*ratio_X
\end{equation}
For example, an integer convolutional kernel $W_i$ is an approximation of its floating-point counterpart $W_f$ by $ratio_W=1/\Delta$. During network inference, these ratios will be accumulated. When the forward propagation is finished, the overall ratio can either be ignored (e.g. for classification) or be reset to 1 by rescaling the output (e.g. for regression). We modify and finetune the FPN network before mapping it to integer network, which do not need any further training.

A low-precision network can be viewed as a high-precision network corrupted by some noise, for example, the noise is introduced by the rounding operation of Equation~(\ref{quantizeW}). During the entire process of integer network inference, there are two kinds of noise: due to weight quantization and due to activation quantization, respectively. Therefore, our work is focused on controlling the quantization noise to ensure the performance of integer networks.\par

\subsection{Activation Quantization With BReLU}
\subsubsection{Why Bounded ReLU}
By discretizing weights before quantizing them into integers, we manage to ensure that integer weights work precisely corresponding to float32 weights. However, integer convolution results in larger numbers. We cannot afford increasing bit-depth after each convolution, thus we need quantize activations from int32 back to int8. Following~\cite{Jacob_2018_CVPR}, we perform activation quantization by multiplication and right shift. Using the symbols in Figure~\ref{fig1}, 
\begin{equation}
V_i=[Y_i*\frac{mul}{2^{shift}}]
\end{equation}
This comes with additional noise due to activation quantization. Different from parameters, activations depend on input data and cannot be trained, which means that activation quantization noise is not removable. For short integers like int8, activation quantization noise is considerably a challenge. For example, our experiment on a 4 layer network~(\ref{vrcnn}) shows that if we use conventional ReLU and always map the largest value into target range without clipping, the performance will drop largely. Besides, dynamically searching for the largest value in a feature map is computationally expensive. Consequently, we propose the activation function with Bounded ReLU, and its adaptability on different datasets and networks. BReLU sets a static range for the activation:
\begin{equation}
f(x)=
\begin{cases}
l,& x\leq l\\
x,&l<x\leq h\\
h,&x>h
\end{cases}
\end{equation}
\begin{figure}[t]
\centering
  \includegraphics[width=1\columnwidth]{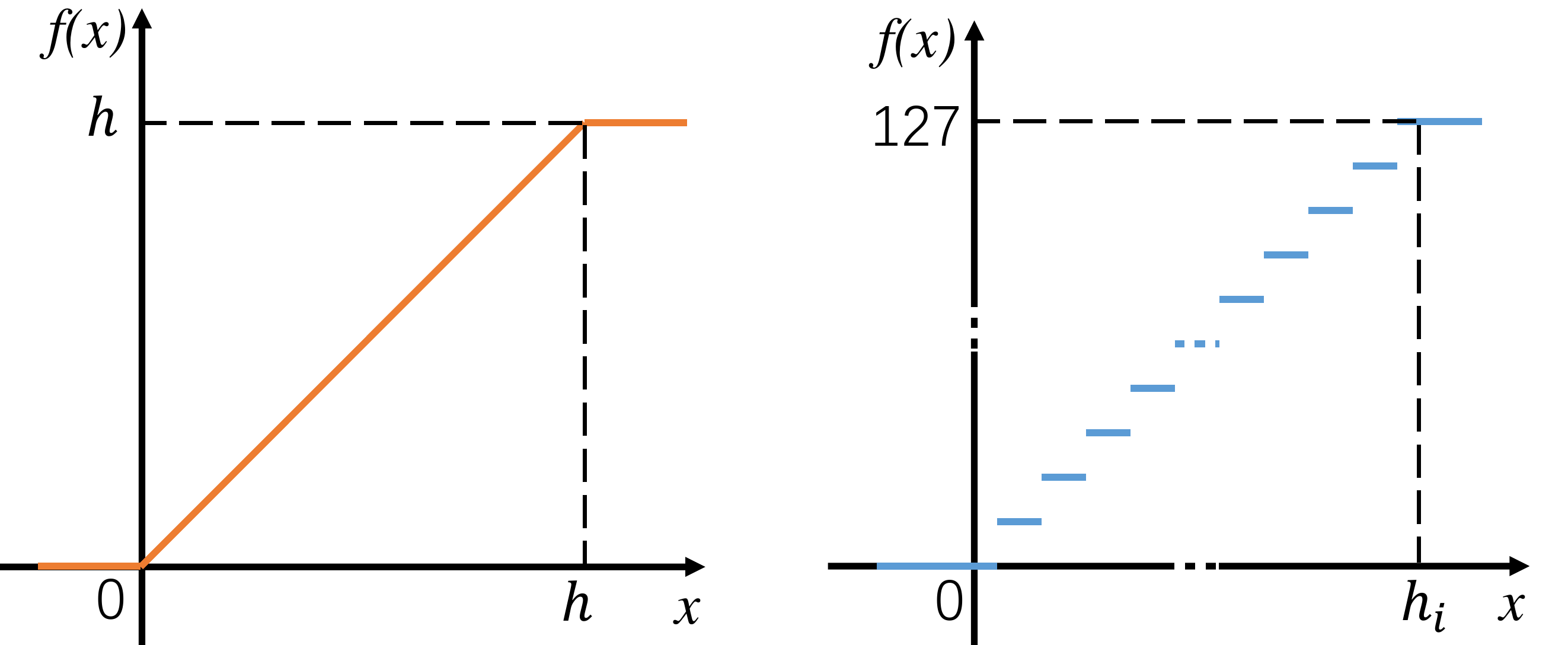}
  \caption{Left: BReLU with $l=0$; Right: Quantized BReLU for 7-bit activation.}
\label{fig2}
\end{figure}

For both FPN networks and integer networks, we use BReLU with $l=0$ (Figure~\ref{fig2}) to replace ReLU. For the upper bound, we have $h_f$ for $Y_f$ and $h_i$ for $Y_i$. To fully exploit the limited integer precision, given an integer data type for $V_i$, $h_i$ is quantized to the maximal possible value $max\_int$ of $V_i$. For example, $max\_int = 127$ for int8. Then we discuss how to set $h_f$. Unlike 32-bit floating-point numbers, short integers have smaller and equally divided representation ranges. Quantizing a float32 feature map into short integer representation is similar to an uniform quantizer~\cite{1094577}, and choosing $h_f$ is actually restricting the range of activations that are ``allowed.'' Clearly, a larger $h_f$ allows more activations to be learned, but also results in a larger quantization step for the uniform quantizer, which will introduce more noise to the output integer representation. Therefore, our methodology is to minimize $h_f$ while keep the learning ability of the floating-point network unaffected, meaning that the network can restore its accuracy after finetuning. 

\subsubsection{Data Domain Adaption}
To achieve minimal $h_f$, it is unavoidable that a group of large values is deactivated among a feature map. Based on the fact that values in a feature map roughly obey a Gaussian distribution with mean and standard deviation being $\mu$ and $\sigma$, we experiment $n$-$\sigma$ rule. $n$-$\sigma$ rule is an empirical rule which provide a range for the vast mojority of a normal distribution. Consequently, we deactivate values which lay beyond the vast mojority. For example, with 3-$\sigma$ rule we set $h_f$ to the minimal value of the largest 0.15\% values in a feature map. $h_f$ is set by computing over the training dataset, and remains unchanged afterwards. Computing over the entire training set, however, is not a good choice because certain outlier samples might take over the largest values. Thus, we calculate $h_f$ in a batch and take the average value over batches as a recommended $h_{rf}$ for float32 network. This value will be further refined in the following. Such method is adaptable for different feature maps generated from different data, and our experiment shows that $n$-$\sigma$ rule is better than ReLU6.

For shallow networks specially, we introduce the Geometric Progression method, which fixes $h_f$ by the training data from scratch and does not need pre-training. Given a network of $n$ layers, the first term $a_0$ and last term $a_n$ are fixed with regard to the range of input and output, and $a_i$ is calculated by a geometric progression:
\begin{equation}
a_i=a_n^{\frac{i}{n}}*a_0^{\frac{n-i}{n}},i=1,2,...,n-1
\end{equation}
$a_i$ is used as $h_{rf}$ for the $i$-th layer. This method provides a BReLU-enabled float32 network from scratch and simplifies the process of network quantization. It works for shallow networks because their learning ability is largely constrained by nature, for which large amounts of deactivated features is acceptable.
\subsubsection{Model Domain Adaption}
$n$-$\sigma$ rule always works fine for a variety of models based on the adaptability of $n$. With different $n$s we can choose different ratios of deactivated values in a feature map. For example, based on 7-bit activations, we find $n=3$ works fine for VDSR and ResNet152, but $n=3.5$ works better for ResNet18. This is because ResNet152 is deeper and require higher precision, thus 3-$\sigma$ rule provide a smaller $h$ and still provide enough learning ability. By experiment, we always try a set of $n$s and for each one apply it to every activation within a network. In theory, the more bits available for activation, the larger $n$ we prefer because it provides stronger learning ability. There are also possibilities that there are different $n$s for different layers among a convolutional neural network, which implicates these layers may play different roles for the final accuracy. We will prove it in the future work.

\subsection{Quantized BReLU}
\label{QBLU}
Beside $h_{rf}$ for floating-point network, we have to quantize it to get its integer counterpart $h_{i}$ (Figure~\ref{fig2}). Take the convolutional layer in Figure~\ref{fig1} for example, we have
\begin{equation}
Y_i \approx Y_f*ratio_Y
\end{equation}
thus
\begin{equation}
h_{ri}=[h_{rf}*ratio_Y]
\end{equation}
we want $h_i$ to be corresponding to $max\_int$, so we can find a pair of two integers $mul$ and $shift$:
\begin{equation}
\label{mul_shift}
mul,shift=mul\text{\_}shift(h_{ri}, max\_int)
\end{equation}
to make sure that
\begin{equation}
max\_int=[h_{ri}*\frac{mul}{2^{shift}}]
\end{equation}
To achieve $ratio_V$ and the following ratios, we shall use $mul$ and $shift$ to rescale $ratio_Y$. However, since $mul$ and $shift$ are integers, it is difficult to ensure that $h_{rf}$ corresponds to $max\_int$ precisely. Thus, we can modify $h_f$. By dequantizing $max\_int$ we have $h_f$ precisely match $max\_int$ by $ratio_V$, as well as all the values clipped by $h_f$. Note that $Y_i$ is not used for convolution, so we simply ensure that $h_i$ is quantized to $max\_int$:

\begin{eqnarray}
ratio_V	&=&	ratio_Y*\frac{mul}{2^{shift}}\\
h_{f}		&=&	max\_int/ratio_V\\
h_{i}		&=&	[h_{f}*ratio_Y]
\end{eqnarray}
Once all the BReLUs in the FPN networks are fixed, we continue to train the float32 network. This will reduce the loss of accuracy for float32 network while minimizing the noise for integer network introduced by activation quantization.
\subsection{Workarounds}
\subsubsection{Re-normalization}
In most cases, input images of a CNN are coded as 8-bit unsigned integer (uint8), which are suitable for integer arithmetic by nature. In TensorRT, the input of 8-bit networks is still kept as floating-point numbers, meaning that the integer inference starts after the first layer. In our integer networks, we simply subtract 128 from the uint8 raw data to get int8 input.\footnote{This is mostly due to a computational issue: convolutional kernel is int8, and it is a bit complex to mix uint8 and int8 in convolution.} Correspondingly, in floating-point networks we also subtract 128 from the raw data and then multiply by $1/ratio_{X}$. This is slightly different from most of the existing networks, which usually normalize the input by subtracting mean and dividing by standard deviation. Fortunately, the different input normalizations seem not influencing the accuracy much, as long as we finetune the networks by original training methods.
\subsubsection{Ratio synchronization between feature maps}
In the forward propagation, the ratios between integer and FPN networks are accumulated along the forward path. Chances are that two paths merge but with different ratios, for example if one path is deeper than the other (e.g. residual connection). In such cases, we need to adjust the ratios of the involved feature maps by rescaling them explicitly or implicitly. We develop Ratio Synchronization between feature maps which ensures the best quantization steps for kernels and same scale for all involved activations. Specific examples will be detailed in the experiments.
\subsubsection{Training}

To recover the accuracy due to quantization, we finetune the network parameters to improve the accuracy similar to \cite{NIPS2016_6573}. Differently, instead of training in the integer domain directly, we would like to finetune the FPN weights, thus preserving the full precision after quantization. First we discretize the weights by the quantization step $\Delta$:
\begin{equation}
W_d=[\frac{W_f}{\Delta}]*\Delta
\end{equation}
Then we use the discretized weights for forward propagation. After backward propagation, we update the original weights by gradients. After finetuning, we quantize the discretized kernel $W_d$, which corresponds to the integer kernel exactly. By avoiding scaling during training stage, our method do not involve pixel-wise division of activations, thus save training time and power. Note that biases are not discretized because they will be quantized to int32 that has enough precision, and the difference between float32 biases and int32 biases can be safely ignored.

As indicated in~\cite{Jacob_2018_CVPR}, a batch normalization (BN) layer rescales the $\Delta$ of the previous convolutional kernel, resulting in different $\Delta$'s for each output channel. Jacob \emph{et al.} \cite{Jacob_2018_CVPR} proposed to merge BN layers into convolutional kernels before training, but failed to recover the precision at last. Similarly, our experiment shows that removing BN before training leads to a significant drop of accuracy. So we use different quantization steps for different channels during training and merge BN parameters into convolution after training.
\subsubsection{Algorithm}
In summary, given a float32 network and training data, we use Algorithm~\ref{algo1} to convert it into an integer network. Our method convert an FPN network into integer network by stages, and each stage is based on the best accuracy of the previous one. Our method require 3 times of finetuning: after re-normalization, after descretization of weights, and after applying BReLU. By applying such method, we hope to find the best quantization step $\Delta$, the best $h_{rf}$, and the best integer model.
\begin{algorithm}[h]
\caption{Convert an FPN network into an integer network}
\label{algo1}
\SetAlgoLined
\SetKwRepeat{Do}{do}{while}
1. Re-normalize the input of the FPN network\;
2. Train the FPN network\;
3. Discretize the weights of the FPN network and record the weight quantization steps\;
4. Train the descretized FPN network\;
5. $n\leftarrow 3$\;
6. \Do{the accuracy is less than step 4 by a predefined threshold}{Calculate the BReLU parameter $h_{rf}$ by the $n$-$\sigma$ rule method\;Calculate the BReLU parameters $h_f$ and $h_i$, and record the activation quantization parameters\;Train the BReLU enabled descretized FPN network\;$n\leftarrow n+0.5$\;}
7. Quantize the weights and biases to get the final integer network.
\end{algorithm}
\section{EXPERIMENTS}
We evaluate our method on state-of-the-art network architectures for both classification and regression tasks. Besides, our method promises bit-wise consistency so we also test on CNNs used in video codec~\cite{10.1007/978-3-319-51811-4_3} which allows only integer computation and demands cross-platform consistency. We use an NVIDIA Titan X (pascal) GPU to measure the computational time.
\subsection{Classification}
Networks used in classification are the backbones of many computer vision tasks such as object detection and semantic segmentation. Among them, ResNet~\cite{He_2016_CVPR} is a widely used network architecture achieving state-of-the-art classification accuracy. It includes multiple residual blocks as well as convolution and pooling layers. Our integer ResNets use only integer arithmetic during inference.

ResNet features many residual connections, for which we develop the ratio synchronization method.
\begin{figure}[t]
  \centering
  \includegraphics[width=0.9\columnwidth]{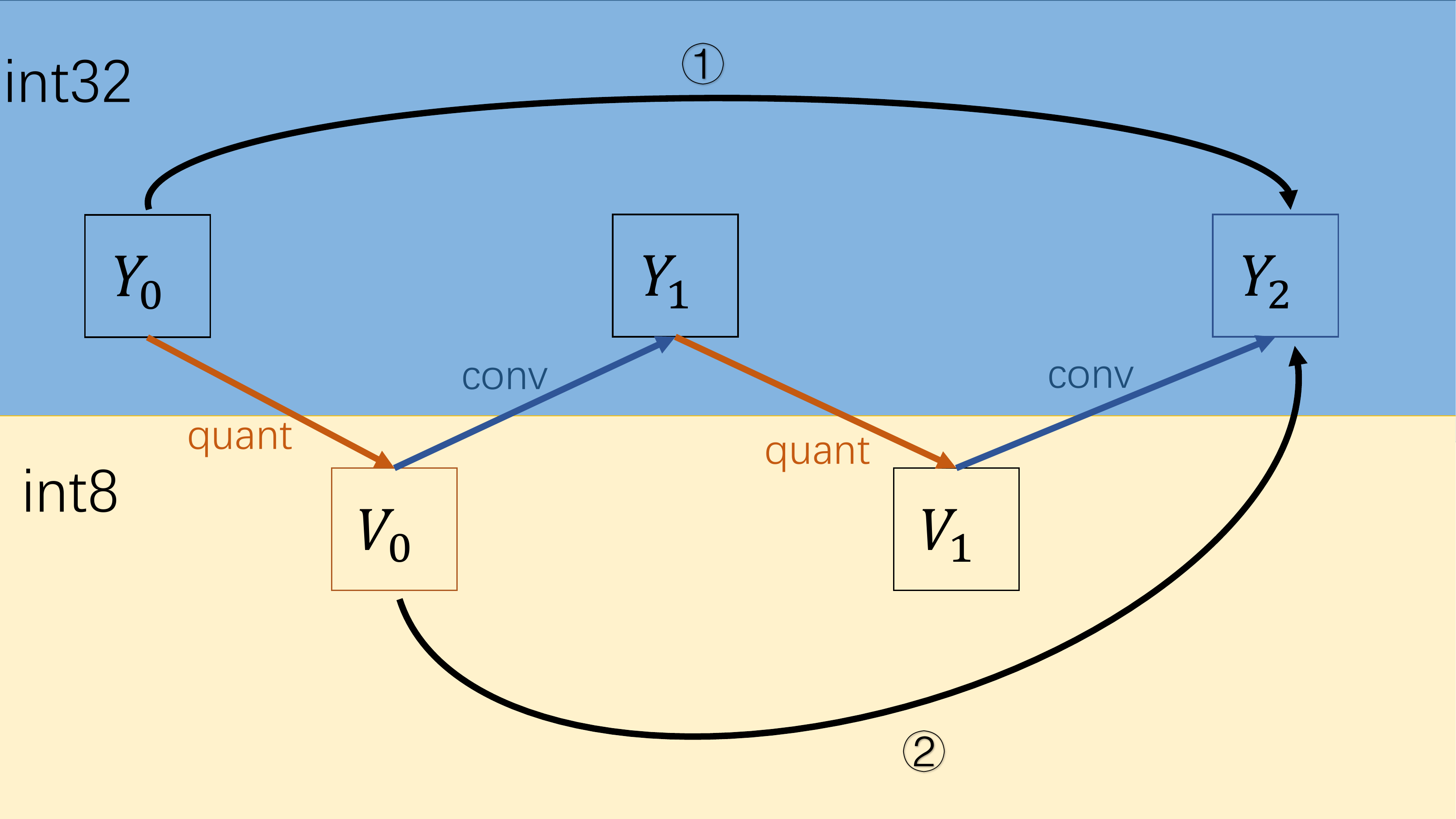}
  \caption{Ratio synchronization for a basic residual block.}
\label{fig3}
\end{figure}
Take basic block in ResNet18 (Figure~\ref{fig3}) for example, the skip connection adds the input of a residual block to its output, directly (type 1) or after convolution (type 2). For type 1, first we rescale $Y_0$ by multiplication and right shift using similar method in Equation~(\ref{mul_shift}) to make the ratio of $Y_0$ close to $Y_2$, then adjust the quantization step $\Delta_2$ of the second convolutional kernel to make the two ratios exactly the same. For type 2, we perform convolution on quantized input $V_0$ and do the similar for the output feature maps. Such procedure is carried out while calculating $h_f$.

By applying Algorithm~\ref{algo1}, we have integer ResNets (Table~\ref{table1}). They achieve equivalent precision with higher speed as well as a quarter model size. Here the model size is the memory footprint of the parameters. For storage purpose, we can surely use compression like Huffman coding to further reduce it. Compared with Jacob et al.~\cite{Jacob_2018_CVPR} which handle activation quantization by exponential moving averages and TensorRT~\cite{migacz20178} which uses relative entropy, our method achieve state-of-the-art performance very close to baseline accuracy. The meaning of this advancement is it serves for very high demanding accuracy, but also provide inference speed up and smaller model size, as shown in Table~\ref{table2}.

\begin{table}
\centering
  \caption{Performance of quantized ResNet on ILSVRC-2012 validation set}
\label{table1}
\begin{threeparttable}
  \begin{tabular}{c c c c c c c}
    \toprule
     Network	& Quantized			& Top-1 (\%)   & Top-5 (\%)   \\
    \midrule
    ResNet18	&float32 			& 69.76  			& 89.08    \\
    ResNet18	&TensorRT 			& 69.56  			& 88.99    \\
    ResNet18	&Ours 			& 70.10  			& 89.36    \\
     \midrule
    ResNet152	&float32			& 78.31  			& 94.06    \\
    ResNet152	&TensorRT			& 74.70  			& 91.78    \\
    ResNet152	&Jacob et al.			& 76.70  			& /\tnote{1}    \\
    ResNet152	&Ours	 			& 78.01  			& 93.81    \\
    \bottomrule
  \end{tabular}
\begin{tablenotes}
\item[1] Not provided in \cite{Jacob_2018_CVPR}.
\end{tablenotes}
\end{threeparttable}
\end{table}

\begin{table}
\centering
  \caption{Model size and inference time of quantized ResNet}
\label{table2}
\begin{threeparttable}
  \begin{tabular}{c c c c c c c}
    \toprule
     Network	& Quantized			& Model size  	& Conv. time\tnote{1}   \\
    \midrule
    ResNet18	&float32 			& 44.6 MB		& 35.1 ms    \\
    ResNet18	&TensorRT 			& 13.3 MB 		& /\tnote{2}    \\
    ResNet18	&Ours 			& 11.1 MB		& 16.2 ms    \\
     \midrule
    ResNet152	&float32			& 230 MB	& 234 ms   \\
    ResNet152	&TensorRT			& 67.6 MB  			& /\tnote{2}    \\
    ResNet152	&Ours	 			& 57.5 MB		& 91.9 ms    \\
    \bottomrule
  \end{tabular}
\begin{tablenotes}
\item[1] Only time for convolutions, average time per batch (50 images).
\item[2] Not comparable because TensorRT adopts other optimizations.
\end{tablenotes}
\end{threeparttable}
\end{table}
To further demostrate the affection of our method including Ratio Syncronization (RS) and BReLU, we compare the result of 4 different configurations in Table~\ref{table3}. However, even we deactivate RS and use simple ReLU6 activation, our result is still significantly better than Jacob et al.~\cite{Jacob_2018_CVPR} and TensorRT. This may due to our whole quantization algorithm and we will open our code and models for researchers.

 Another finding is that within a single convolutional kernel with batch normalization, there are different $\Delta$'s for different channels, and some of them is extremely large that the whole channel after quantization is 0. Based on the experiment that the accuracy is not affected, we believe that these channels can be safely pruned, which will further reduce model size and speed up inference.
\begin{table}
  \centering
  \caption{Performance of integer ResNet152 with different configurations}
\label{table3}
     \begin{tabular}{c c c}
    \toprule
    configuration					& Top-1 (\%)   & Top-5 (\%) \\
    \midrule
FPN					& 78.314&94.060  \\  
ReLU6					& 77.602&93.650 \\  
ReLU6+RS				& 77.810&93.718 \\  
BReLU					& 77.754&93.656 \\  
BReLU+RS				& 78.008&93.806\\
    \bottomrule
  \end{tabular}
\end{table}
\subsection{Regression}
A lot of image processing tasks require regression CNNs, such as image super-resolution, denoising, and generative models. Nonetheless, low-precision networks for regression tasks are seldom reported before. Since regression networks do not have average pooling or non-maximum suppression, they are less tolerant to the noise during inference. By applying our method on multiple regression tasks, we show that our method is also qualified for regression networks.
\subsubsection{VDSR for image super-resolution}
Very deep convolutional networks for image super-resolution (VDSR)~\cite{Kim_2016_CVPR} is a typical CNN architecture used for regressive tasks, constructed by 20 convolutional layers with
kernel size 3$\times$3. The output is an image of same size with input, but looks better as if it is ``super-resolved.'' VDSR is widely used in end-to-end image tasks like image-coloring, image-denoising or image-deraining. Our result is shown in Table~\ref{table4}. Beside 7-bit activation, we also evaluated our method on down to 4-bit activations. As shown in Table~\ref{table5}, the accuracy drops significantly from 5-bit to 4-bit. Generally speaking, the deeper or less tolerant to noise a network is, the higher precision for activations is needed.\par
\begin{figure}[h]
  \centering
  \includegraphics[width=0.9\columnwidth]{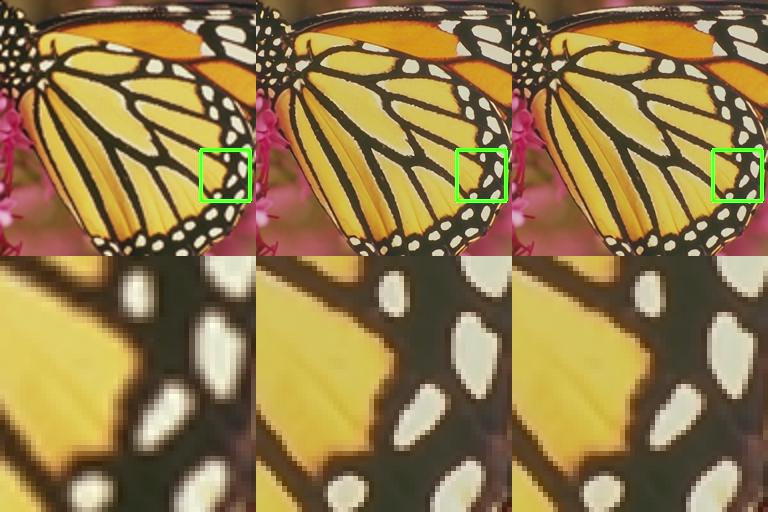}
  \caption{Left: Bicubic interpolated image; Middle: SR result by our int8 network; Right: SR result by the original float32 network.}
\label{fig4}
\end{figure}
Single image super-resolution is widely used in computer vision applications ranging from security and surveillance imaging to medical imaging where more image details are required on demand. Therefore, it not only requires objective evaluation, but also subjective quality. Our result in Figure~\ref{fig4} demonstrate that output of our integer network is both accurate by PSNR and visual pleasing as well as output of float32 network.
\begin{table}
\centering
  \caption{Performance of quantized VDSR on Set-5}
\label{table4}
\begin{threeparttable}
  \begin{tabular}{c c c c c c c c c}
    \toprule
     	& float32	& int8 \\
    \midrule
  PSNR for 2$\times$	& 37.55  		& 37.51		\\
	PSNR for 3$\times$	& 33.78  		& 33.78		\\
	PSNR for 4$\times$	& 31.44  		& 31.42		\\
	Model size	(MB)	& 2.54  	& 0.65 		\\
	Conv. time (ms)\tnote{1}& 120  	& 55.7 		\\
    \bottomrule
  \end{tabular}
\begin{tablenotes}
\item[1] Only time for convolutions, overall time of Set-5.
\end{tablenotes}
\end{threeparttable}
\end{table}

\begin{table}
  \caption{Performance of quantized VDSR with 8-bit weights and activations of different bit-depths}
\label{table5}
  \centering
  \begin{tabular}{c c c c c}
    \toprule
     Bit-depth  & 2$\times$ PSNR   & 3$\times$ PSNR        & 4$\times$ PSNR\\
    \midrule
    8   & 37.52     & 33.78       & 31.42\\
    7   & 37.51     & 33.78       & 31.42\\
    6   & 37.49     & 33.76       & 31.40\\
    5   & 37.39     & 33.69       & 31.29\\
    4   & 36.08     & 33.24       & 30.74\\
    \midrule
    float32   & 37.55     & 33.78       & 31.44\\
    \bottomrule
  \end{tabular}
\end{table}
\subsubsection{VRCNN for compression artifact reduction}\label{vrcnn}
VRCNN~\cite{10.1007/978-3-319-51811-4_3} is a network used in video coding for compression artifact reduction. It works similarly with VDSR, consisting of 4 convolution layers. VRCNN has 2 concatenation layers, each of which has two kinds of kernel sizes, like in Inception Net~\cite{Szegedy_2015_CVPR}. We design the ratio synchronization method for concatenation layers, as shown in Algorithm~\ref{algo2}. To scale all the involved feature maps into same ratio, we select the minimal ratio as reference to ensure that none feature map would overflow. Then we adjust each quanization parameter $mul$ and $shift$ to reach that ratio, as well as corresponding $\Delta$ to promise precise syncronization. We perform similar process on skip connections in ResNets above.

\begin{algorithm}[h]
 \caption{Ratio synchronization for concatenation layers}
\label{algo2}
\SetAlgoLined
\SetKwInOut{Input}{input}\SetKwInOut{Output}{output}
\Input{$\Delta_i, ratio_{Y_i}, ratio_{V_i}, i=1,2,...,N$}
\Output{$\Delta_i, ratio_{Y_i}, mul_i, shift_i, ratio_{V_i}, i=1,2,...,N$}
\BlankLine
$ratio\_min_{V}=\min(ratio_{V_i})$\;
\For{$i\gets1$ \KwTo $N$ }{
    $ratio_{V_i}=ratio\_min_{V}$\;
    $mul_i,shift_i=mul\_shift(ratio_{Y_i},ratio_{V_i})$\;
    $\Delta_i=ratio_{Y_i}*mul_i/(2^{shift_i}*ratio_{V_i})$\;
    $ratio_{Y_i}=ratio_{V_i}*2^{shift_i}/mul_i$\;
    }
\end{algorithm}
VRCNN is a shallow network, for which we calculated $h_f$ by the simple Geometric Progression method. Specifically, we set $a_0$ to 0.5, which is the absolute max value of input after re-normalization, and $a_4$ to the max absolute value of the output residual. We train the BReLU-FPN network from scratch and quantize the BReLU-FPN network, leading to integer VRCNN with same performance as its float32 couterpart.

\begin{table}
\centering
  \caption{Performance of quantized VRCNN on Classes A--E}
\label{table6}
\begin{threeparttable}
  \begin{tabular}{c c c c c c}
    \toprule
      	& BD-rate	& Model size (KB) 	& Inf. time (ms)\tnote{1}  \\
    \midrule
    float32 	& $-$5.3\%  			& 214    		& 155	\\
    int8 		& $-$5.3\%  			& 54    		& 79		\\
    \bottomrule
  \end{tabular}
\begin{tablenotes}
\item[1] Entire inference time for a single picture of size 2560$\times$1600.
\end{tablenotes}
\end{threeparttable}
\end{table}
Performance of VRCNN is evaluated on standard HEVC test sequences with QP ranging from 22 to 37, as shown in Table~\ref{table6}. Note that cross-platform consistency is crucial in video codec, which becomes one of the motivations of our work.

\section{CONCLUSION}
We have proposed a method for converting FPN networks into integer-arithmetic-only networks without sacrificing accuracy. Our key idea is to replace ReLU with BReLU and adapt it to various datasets and networks, so as to efficiently quantize the activations. For the upper bound of BReLU we have studied two methods for shallow and deep networks respectively.

We have tested our methods on three tasks, including image super-resolution and compression artifact reduction that are not reported before. Our method can achieve efficient integer networks that have a quarter model size and double speed on modern GPUs, and ensure cross-platform consistency. Our method outperforms Jacob et al.~\cite{Jacob_2018_CVPR} and TensorRT in accuracy and achieves state-of-the-art performance.

In the future, we plan to extend our method for other CNNs that have customized or specialized units, such as leaky ReLU and sigmoid (e.g. in LSTM). We will also conduct further study into BReLU by applying different stratagy for different layers.

\bibliography{ecai}
\end{document}